\documentclass{vgtc}                         
\usepackage{color}

\ifpdf
  \pdfoutput=1\relax                   
  \usepackage{graphicx}                
  \DeclareGraphicsExtensions{.pdf,.png,.jpg,.jpeg} 
\else
  \usepackage{graphicx}                
  \DeclareGraphicsExtensions{.eps}     
\fi%

\graphicspath{{figures/}{pictures/}{images/}{./}} 

\usepackage{microtype}                 
\PassOptionsToPackage{warn}{textcomp}  
\usepackage{textcomp}                  
\usepackage{mathptmx}                  
\usepackage{times}                     
\usepackage{cite}                      
\usepackage{tabu}                      
\usepackage{booktabs}                  

\usepackage[draft]{hyperref}

\onlineid{1094}

\vgtccategory{Research}

\vgtcinsertpkg

\title{3D Virtual Garment Modeling from RGB Images}

\author{Yi Xu\thanks{e-mail: yi.xu@oppo.com, currently with OPPO US Research Center. The work was done when Yi Xu was with JD.} \\ %
        \scriptsize OPPO US Research Center%
\and Shanglin Yang\thanks{e-mail:shanglin.yang@jd.com} \\ %
        \scriptsize JD.COM American Technologies Corporation %
\and Wei Sun\thanks{e-mail:wsun12@ncsu.edu, the work was when Wei Sun was with JD.} \\ %
        \scriptsize North Carolina State University %
\and Li Tan \thanks{e-mail:tanli5@jd.com} \\ %
        \scriptsize JD.COM %
\and Kefeng Li\thanks{e-mail:likefeng@jd.com} \\ %
        \scriptsize JD.COM %
\and Hui Zhou\thanks{e-mail: hui.zhou@jd.com} \\
        \scriptsize \centering JD.COM American Technologies Corporation} %

\abstract{We present a novel approach that constructs 3D virtual garment models from photos. Unlike previous methods that require photos of a garment on a human model or a mannequin, our approach can work with various states of the garment: on a model, on a mannequin, or on a flat surface. To construct a complete 3D virtual model, our approach only requires two images as input, one front view and one back view. We first apply a multi-task learning network called JFNet that jointly predicts fashion landmarks and parses a garment image into semantic parts. The predicted landmarks are used for estimating sizing information of the garment. Then, a template garment mesh is deformed based on the sizing information to generate the final 3D model. The semantic parts are utilized for extracting color textures from input images. The results of our approach can be used in various Virtual Reality and Mixed Reality applications.
} 

\CCScatlist{
  \CCScatTwelve{Computing methodologies}{Computer graphics}{Graphics systems and interfaces}{};
  \CCScatTwelve{Computing methodologies}{Artificial intelligence}{Computer vision}{Computer vision problems}{};
}

\begin{document}



\maketitle

\section{Introduction} 
Building 3D models of fashion items has many applications in Virtual Reality, Mixed Reality, and Computed-Aided Design (CAD) for apparel industry. A lot of commercial efforts have been put into this field. For example, there are a few CAD software systems that are created for 3D garment design, but most of them focus on creating 3D garment models based on 2D sewing patterns, such as MavelousDesigner and Optitex. Recently, a few e-commerce platforms have begun to use 3D virtual garments to enhance online shopping experiences. However, large variation, short fashion product life cycle, and high modeling costs make it difficult to use virtual garments on a regular basis. This necessitates a simple yet effective approach for 3D garment modeling. 

There have been a lot of research for creating 3D virtual garment models. Some use specialized multi-camera setups to capture 4D evolving shape of the garments \cite{Bradley:2008:MGC:1360612.1360698, Pons-Moll:2017:CSC:3072959.3073711}. These setups are complicated; therefore limiting their usage. Other methods take 2D sewing patterns \cite{Berthouzoz:2013:PSP:2461912.2461975} or 2D sketches \cite{Robson:2011:SFP:1994025.1994424} as input and build 3D models that can be easily manufactured. Although these methods use 2D images as input, they still rely on the careful and lengthy design of expert users. Another group of methods deform/reshape 3D template meshes to design garments that best fit 3D digital human models  \cite{Meng:2012:FSC:2076818.2077262}. This can be an overkill in certain applications where an accurate design is not needed. Recently, there have been some methods that create 3D garment models from a single image or a pair of images \cite{Danzrek:2017:DGS:3128975.3129001, Jeong:2015:doi:10.1002/cav.1653, Yang:2018:TBD, Zhou:2013:doi:10.1111/cgf.12215}. All of these methods assume the garment is worn by a human model or a mannequin; therefore, do not provide the convenience of working with readily available photos.

We propose a method that can construct 3D virtual garment models from photos that are available on the web, especially on e-commerce sites. Fig. \ref{fig_input} shows two examples. Each photo set displays several different views of a piece of garment on a fashion model, on a mannequin, or flattened on a support surface. To generate a 3D virtual model, a user needs to specify one front and one back image of the garment. The generated 3D model is up to a scale, but can have absolute scale if user specifies a real world measurement (e.g., sleeve length in meters). 

We train a multi-task learning network, called JFNet, to predict fashion landmarks and segment a garment image into semantic parts (i.e., left sleeve, front piece, etc.). Based on the landmark predictions, we estimate sizing information of the garment and deform a template mesh to match the estimated measurements. We then deform the semantic parts onto a 2D reference texture to lift textures. It is worth-noting that our method is capable of using a single image as input if front-back symmetry is assumed for a garment. Our contributions are as follows:

\begin{figure}[t!]
\begin{center}
   \includegraphics[width=0.98\linewidth]{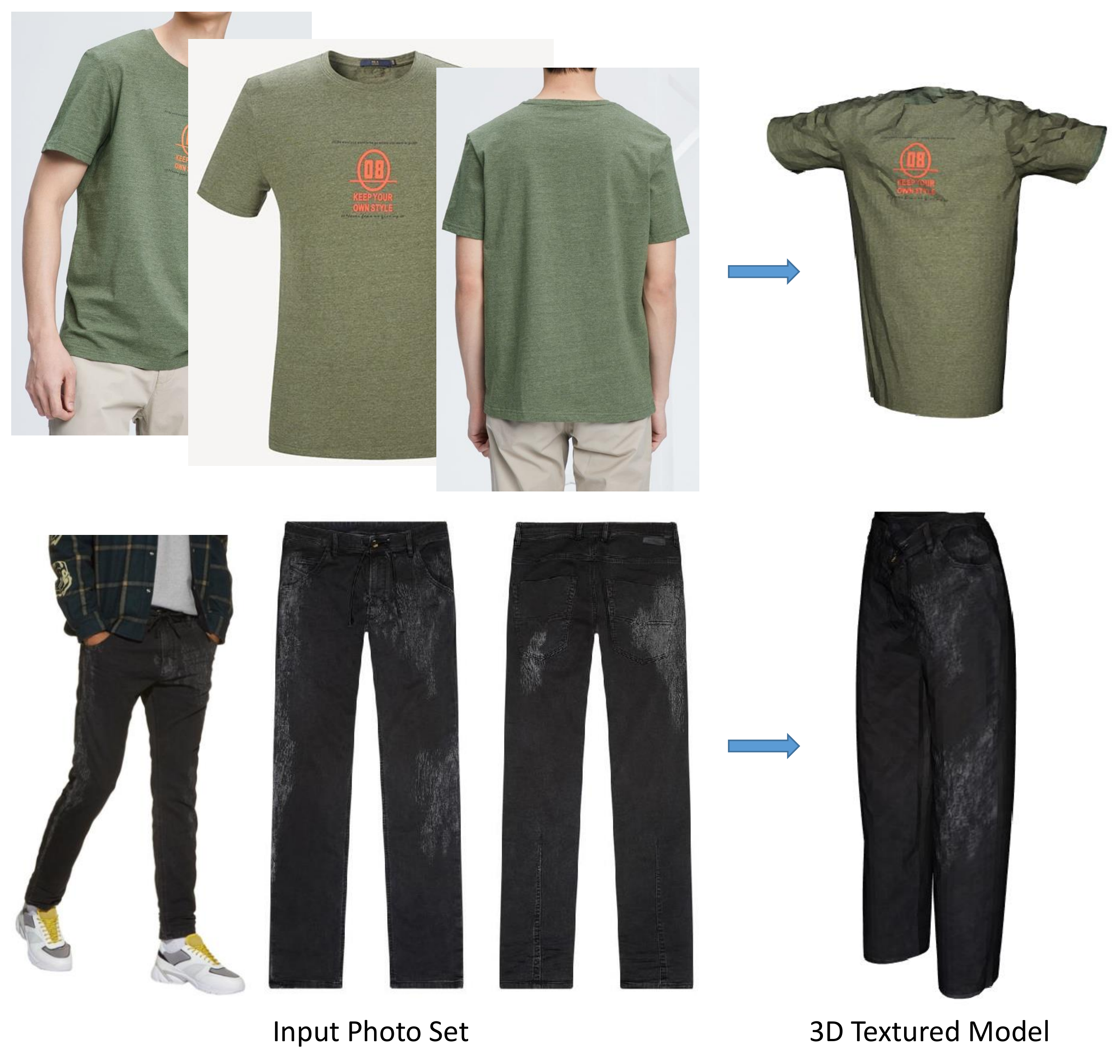}
   \caption{Two product photo sets (left) on an e-commerce site and 3D textured models (right) computed using two photos from each input set.}
\label{fig_input}
\end{center}
\vspace{-15pt}
\end{figure}

\begin{itemize}
    \item We present a complete and easy-to-use approach that generates a 3D textured garment model using product photo set. T-shirt and pants are modeled in this paper; however, our approach can be extended to other garment types.
    
    \item We propose a multi-task learning framework that predicts fashion landmarks and segments garment image into semantic parts.
    
    \item We present algorithms for size estimation and texture extraction from garment images.
\end{itemize}

\begin{figure*}[h!]
\begin{center}
   \includegraphics[width=0.99\linewidth]{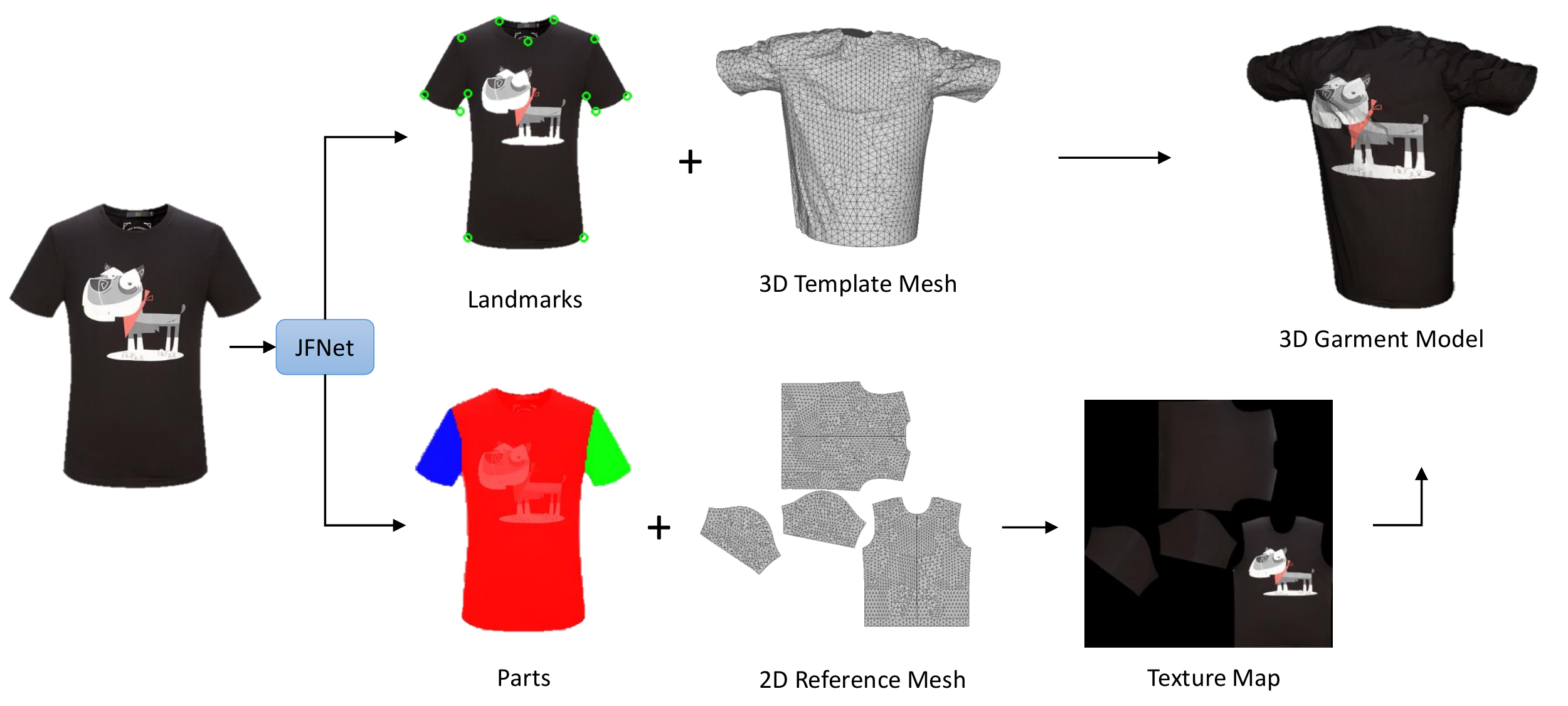}
   \caption{System Overview. For each input image, we jointly predict landmark locations and segment the garment into semantic parts using the proposed JFNet. The predicted landmarks are used to guide the deformation of a 3D template mesh. The segmented parts are used to extract garment textures. Finally, 3D textured garment model is produced.}
\label{fig_overview}
\end{center}
\end{figure*}

\section{Related Work}
In this section, we discuss related work in garment modeling, joint human body and garment shape estimation, semantic parsing of fashion images, and image-based virtual try-on.

\subsection{Garment Modeling and Capturing}
Garment modeling methods can be classified into the following three categories: geometric approaches, image-based 3D reconstruction, and image-based template reshaping. 

\subsubsection{Geometric Approaches}
Methods in this category typically have roots from the CAD community. Wang et al. \cite{Wang:2005:DAC:1649587.1649870} automated the Made-to-Measure (MtM) process by fitting 3D feature templates of garments onto different body shapes. Meng et al. \cite{Meng:2012:FSC:2076818.2077262} proposed a method that preserves the shape of user-defined features on the apparel products during the automatic MtM process.

Other methods use 2D sketches or patterns as input. For example, Decaudin et al. \cite{Decaudin:2006:doi:10.1111/j.1467-8659.2006.00982.x} fitted garment panels to contours and seam-lines that are sketched around a virtual mannequin. These panels are then approximated with developable surfaces for garment manufacturing.  Robson et al.  \cite{Robson:2011:SFP:1994025.1994424} created 3D garments that are suitable for virtual environments form simple user sketches using context-aware sketch interpretation. Berthouzoz et al. \cite{Berthouzoz:2013:PSP:2461912.2461975} proposed an approach that parses existing sewing patterns and converts them into 3D models. Wang et al. \cite{garmentdesign_Wang_SA18} presented a system that is capable of estimating garment and body shape parameters interactively using a learning approach. All of these methods rely on certain level of tailoring expertise from users.

\subsubsection{Image-based 3D Reconstruction}
Some approaches aimed to create 3D models directly from input images and/or videos of a garment. Early work by White et al. \cite{White:2007:CAO:1276377.1276420} used a custom set of color markers printed on the cloth surface to recover 3D mesh of dynamic cloth with consistent connectivity. Markerless approaches were also developed by using multi-camera setup \cite{Bradley:2008:MGC:1360612.1360698}, multi-view 3D scans with active stereo\cite{Pons-Moll:2017:CSC:3072959.3073711}, or depth cameras \cite{Chen:2015:GMD:2816795.2818059}. These methods require specialized hardware and do not work with existing garment photos.

\subsubsection{Shape Parameter Estimation}
Our approach is most similar to methods that utilize parametric models of human and/or garments. Zhou et al. \cite{Zhou:2013:doi:10.1111/cgf.12215} took a single image of a human wearing a garment as input. Their approach first estimates human pose and shape from images using parameter reshaping. Then, a semi-automatic approach is used to create an initial 3D mesh for the garment. Finally, shape-from-shading is used to recover details. Their method requires user input for pose estimation and garment outline labeling, assumes the garment is front-back symmetric, and does not extract textures from the input image.

Jeong et al. \cite{Jeong:2015:doi:10.1002/cav.1653} fitted parameterized pattern drafts to input images by analyzing silhouettes. However, their method requires input images of a mannequin both with and without garment from the same viewpoint. Yang et al. \cite{Yang:2018:TBD} used semi-automatic processing to extract semantic information from a single image of a model wearing the garment and used optimization with a physics-inspired objective function to estimate garment parameters. Compared to this method, our method provides a more advanced joint learning model for semantic parsing.

The DeepGarment framework proposed by Dan\'{z}\u{r}ek et al. \cite{Danzrek:2017:DGS:3128975.3129001} learns a mapping from garment images to 3D model using  Convolutional Neural Networks (CNN). More specifically, the learned network can predict displacements of vertices from a template mesh. However, garment texture is not learned.

\subsection{Joint Human Body and Garment Shape Estimation}
There have been a lot of efforts that address the challenging problem of joint human body and garment shape estimation. 

Alldieck et al. \cite{video_based_people_model_cvpr2018} reconstructed detailed shape and texture of clothed human by transforming a large amount of dynamic human silhouettes from a single RGB sequence to a common reference frame. Later, the same authors introduced a learning approach that only requires a few RGB frames as input \cite{Octopus_alldieck19cvpr}.
Natsume et al. \cite{SiCloPe_2019} reconstructed a complete and textured 3D model of a clothed person using just one image. In their work, deep visual hull algorithm is used to predict 3D shape from silhouettes and a Generative Adversarial Network (GAN) is used to infer the appearance of the back of the human subject. Habermann et al. \cite{LiveCap_2019} presented a system for real time tracking of human performance, but relied on a personalized and textured 3D model that was captured during a pre-processing step. These work do not separate underlying body shape from garment geometry.

Using RGBD camera as input device, body shape and garment shape can be separated. For example, Zhang et al. \cite{Clothed_3D_scan_2017} reconstructed naked human shape under clothing. Yu et al. \cite{DoubleFusion_Yu_2018_CVPR} used a double layer representation to reconstruct geometry of both body and clothing. Physics based cloth simulation can also be incorporated into the framework to better track human performance \cite{SimulCap_Yu_2018_CVPR}.

\subsection{Fashion Semantic Parsing}
In this section, we review related work in fashion landmark prediction, semantic segmentation, and multi-task learning.

\subsubsection{Fashion Landmark Prediction}
Fashion landmark prediction is a structured prediction problem for detecting functional key points, such as corners of cuff, collar, etc. Despite it being a relatively new topic \cite{Liu:2016:fashionlandmark}, it has roots in a related problem-human pose estimation. Early work on human pose estimation used pictorial structures to model spatial correlation between human body parts \cite{Andriluka:2009:PSR}. Such method only works well when all body parts are visible, so that the structure can be modeled by graphical models. Later on, hierarchical models were used to model part relationships at multiple scales \cite{Tian:2012:SHM}. Spatial relationship can also be learned implicitly using a sequential prediction framework, such as Pose Machines \cite{Ramakrishna:2014:PM}. CNNs can also be integrated into Pose Machines to jointly learn image features and spatial context features \cite{Wei:2016:CPM}. 

Different from human pose, fashion landmark detection predicts functional key points of fashion items. Liu et al. proposed a Deep Fashion Alignment (DFA) \cite{Liu:2016:fashionlandmark} framework that cascades CNNs in three stages similar to DeepPose \cite{Toshev:2014:DP}. To achieve scale invariance and remove background clutter, DFA assumes that bounding boxes are known during training and testing; thus limiting its usage. This constraint was later removed in Deep LAndmark Network (DLAN) \cite{Yan:2017:UFL:3123266.3123276}. It is worth noting that the  landmarks defined in these approaches cannot be used for texture extraction. For example, a mid-point on the cuff is a landmark defined in their work. In our work, two corners of the cuff are predicted and they carry critical information for texture extraction.

\subsubsection{Semantic Segmentation}
Semantic segmentation assigns semantic labels to each pixel. CNNs have been successfully applied to this task. Long et al. proposed Fully Convolutional Networks (FCNs)  for semantic segmentation \cite{Long:2015:FCN}, which achieved significant improvements over methods relied on hand-crafted features. Built upon FCNs, Encoder-Decoder architectures have shown great success  \cite{Ronneberger:2015:UNET:10.1007/978-3-319-24574-4_28, Badrinarayanan:2017:SEGNET}. Such an architecture typically has an encoder that reduces feature map and a decoder that maps the encoded information back to input resolution. Spatial Pyramid Pooling (SPP) can also be applied at several scales to leverage multi-scale information \cite{Zhao:2017:PSP}. DeepLabV3+  \cite{Chen:2018:DEEPLABV3+} combines the benefits of both SPP and Encoder-Decoder architecture to achieve state-of-the-art result. Our part segmentation sub-network is based on DeepLabV3+ architecture.
Similar to our work, Alldieck et al. \cite{Detail_human_avartar_2018} also used human semantic part segmentation to extract detailed textures from RGB sequences.

\subsubsection{Multi-task Learning}
Multi-task learning (MTL) has been used successfully for many applications due to the inductive bias it achieves when training a model to perform multiple tasks. Recently, it has been applied to several computer vision tasks.  Kokkinos introduced UberNet \cite{Kokkinos:2018:UBN} that can jointly handle multiple computer vision tasks, ranging from semantic segmentation, human parts, to object detection. Ranjan et al. proposed HyperFace \cite{Ranjan:2018:HYPERFACE} for simultaneously detecting faces, localizing landmarks, estimating head pose, and identifying gender. Perhaps the most similar work to ours is the work of JPPNet \cite{Liang:2018:LIP}. It is a joint human parsing and pose estimation network, while our work uses MTL for garment image analysis.  
Another MTL work on human parsing from the same group is 
\cite{Human_parsing_Gong_2018_ECCV}, where semantic
part segmentation and instance-aware edge detection are jointly learned. 

\subsection{Image-based Virtual Try-on}
As an alternative to 3D modeling, image-based virtual try-on has also been explored. Neverova et al.  \cite{DensePoseTransfer_Neverova_2018_ECCV} used a two-stream network where a data-driven predicted image and a surface-based warped image are combined and the whole network is learned end-to-end to generate a new pose of a person. Lassner et al. \cite{Lassner_2017_ICCV} used only image information to predict images of new people in different clothing items. VITON \cite{VITON_2018} on the other hand transfers the image of a new garment onto a photo of a person. 


\section{Our Approach}
In this section, we explain our approaches on garment image parsing, 3D model creation, and texture extraction. Fig. \ref{fig_overview} shows an overview of our approach.

\begin{figure}[bh!]
\begin{center}
\includegraphics[width=0.90\linewidth]{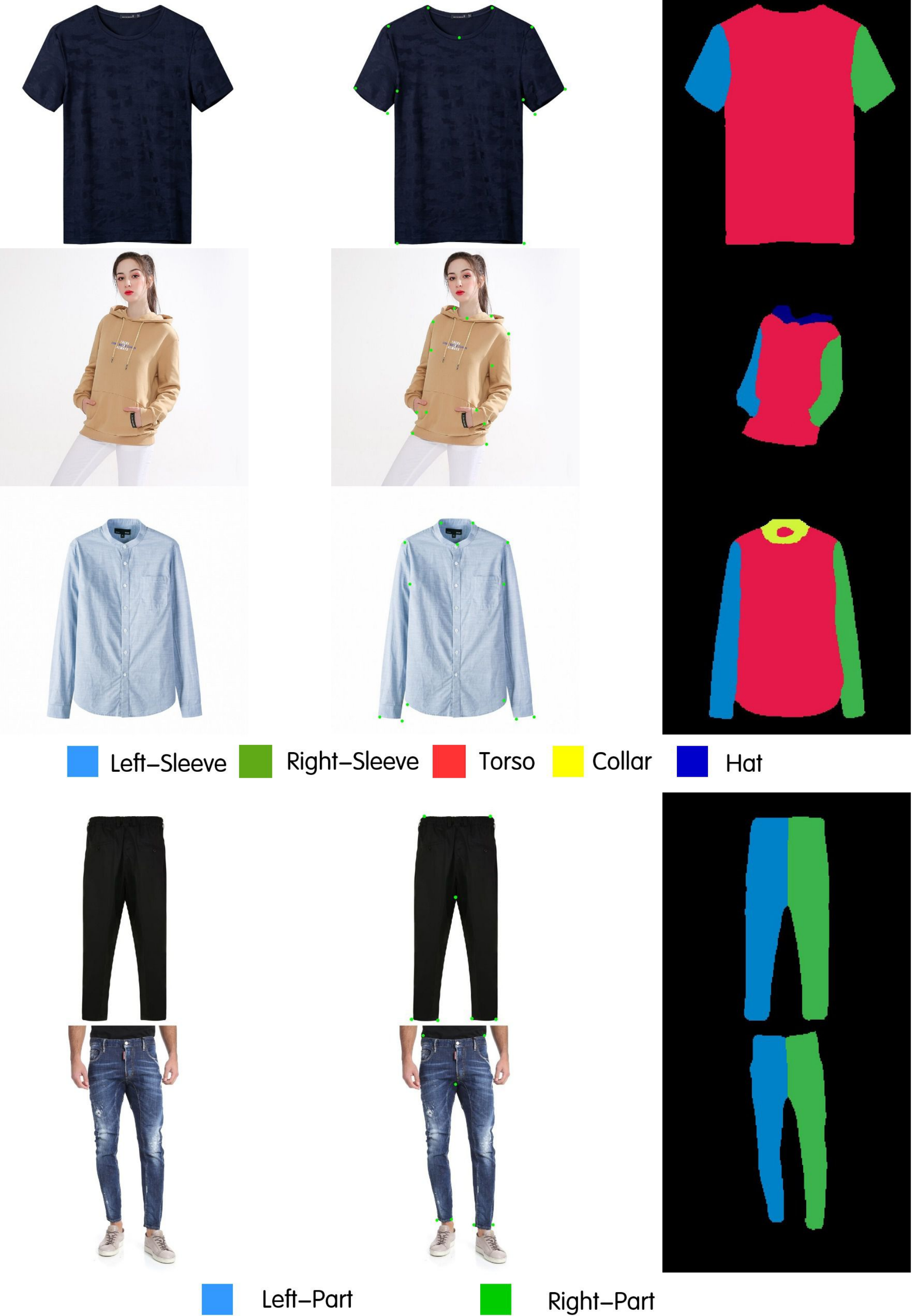}
\caption{Annotation Examples. Top and bottom shows landmark and part labeling for tops (including T-shirt) and pants respectively. }
\label{fig_data}
\end{center}
\end{figure}

\begin{figure*}[h!]
\begin{center}
   \includegraphics[width=0.95\linewidth]{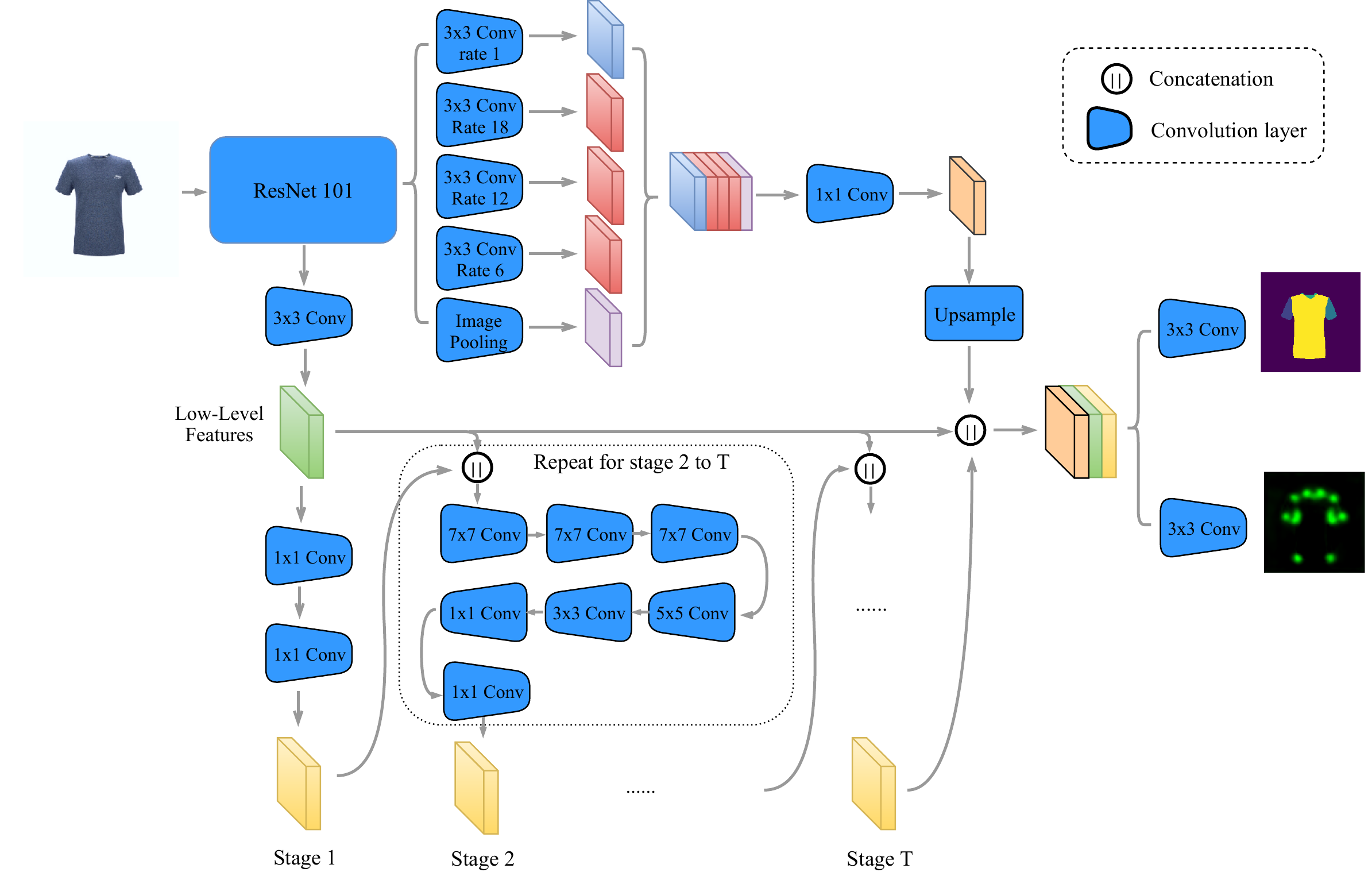}
\end{center}
   \caption{JFNet. Our proposed multi-task learning model use ResNet-101 as backbone network to extract shared low level features. For landmark prediction (bottom half), we apply \textit{T}-stage CNNs. Each stage refines the prediction iteratively. For garment part segmentation, Atrous Spatial Pyramid Pool (ASPP) is applied on the ResNet output and followed by 1x1 convolution and up-sampling. At the last stage of the network, results from two branches are concatenated together for joint learning. 
   }

\label{fig_jfnet}
\end{figure*}

\subsection{Data Annotation}
To train JFNet, we built a dataset with both fashion landmarks and pixel-level segmentation annotations. We collected 3,000 images of tops (including T-shirts) and another 3,000 images of pants from the web. For each type of garment, a set of landmarks are defined based on fashion design. 13 landmarks are defined for tops including center and corners of neckline, corners of both cuffs, end points on hemline, and armpits. 7 landmarks are defined for pants including end points of waistband, crotch, and end points of the bottom. 

For part segmentation, we defined a set of labels and asked the annotators to provide pixel-level labeling. For tops, we used 5 labels including left-sleeve, right-sleeve, collar, torso, and hat. For pants, we used 2 labels including left-part and right-part. Some labeling examples are shown in Fig. \ref{fig_data}.

\subsection{Garment Image Parsing}
Our joint garment parsing network JFNet built upon Convoluitional Pose Machines (CPMs) \cite{Wei:2016:CPM} for landmark prediction and DeepLabV3+ \cite{Chen:2018:DEEPLABV3+} for semantic segmentation.

The network architecture of JFNet is illustrated in Fig. \ref{fig_jfnet}. We use ResNet-101 \cite{He:2016:ResNet} as our backbone network to extract low-level features. Then we use two branching networks to obtain landmark prediction and part segmentation. Finally, we use a refinement network to refine the prediction results.

\subsubsection{Landmark Prediction}

For landmark prediction (bottom half of Fig. \ref{fig_jfnet}), we use a learning network with \textit{T}-stages similar to that of \cite{Wei:2016:CPM}. At first stage, we extract second stage outputs of ResNet-101 (Res-2) followed by a 3x3 convolutional layer as low level features from the input image. Then, we use two 1x1 convolutional layers to predict landmark heatmap at the first stage. At each of the subsequent stages, we concatenate the landmark heatmap predicted from the previous stage with shared low-level features from Res-2. Then we use five convolutional layers followed by two 1x1 convolutional layers to predict the heatmap at the current stage. The architecture repeats this process for \textit{T} stages, where the size of receptive field increases with each stage. This is crucial for learning long-range relationships between fashion landmarks. The heatmap at each stage is compared against labeled ground truth and calculated towards total training loss.

\subsubsection{Garment Part Segmentation}
For semantic garment part segmentation (top half of Fig. \ref{fig_jfnet}), we followed the encoder architecture of DeepLabV3+ \cite{Chen:2018:DEEPLABV3+}. Atrous Spatial Pyramid Pool (ASPP) module, which can learn context information at multiple scales effectively, is applied after the last stage output of ResNet-101, followed by one 1x1 convolutional layer and up-sampling. 

\subsubsection{Refinement}
To refine landmark prediction and part segmentation, and to promote each other, we concatenate the landmark prediction result from the \textit{T}-th stage of the landmark sub-network, the part segmentation result from the segmentation sub-network, and the shared low-level features together. We then apply a 3x3 convolutional layer for landmark prediction and part segmentation respectively. The sum of loss from both branches is used for jointly training the network end-to-end.

\subsubsection{Training Details}
We load ResNet-101 parameters that are pre-trained on ImageNet classification task. During training, random crop and random rotation between -10 and 10 degrees are applied for data augmentation and the final input image size is resized to 256x256. We adopt SGD optimizer with 0.9 as momentum. Learning rate is initially set as 0.001 and ``poly" decay \cite{Zhao:2017:PSP} is set to $10^{-6}$ in 100 total training epoches.

\subsection{3D Model Construction}
Our approach uses fashion landmarks to estimate the sizing information and to guide the deformation of a template mesh. Textures are extracted form input images and mapped onto the 3D garment model. In this section, we first discuss the garment templates used in our system. Then, we discuss our 3D modeling and texturing approaches.

\subsubsection{Garment Templates}

\begin{figure}[t!]
\begin{center}
   \includegraphics[width=0.95\linewidth]{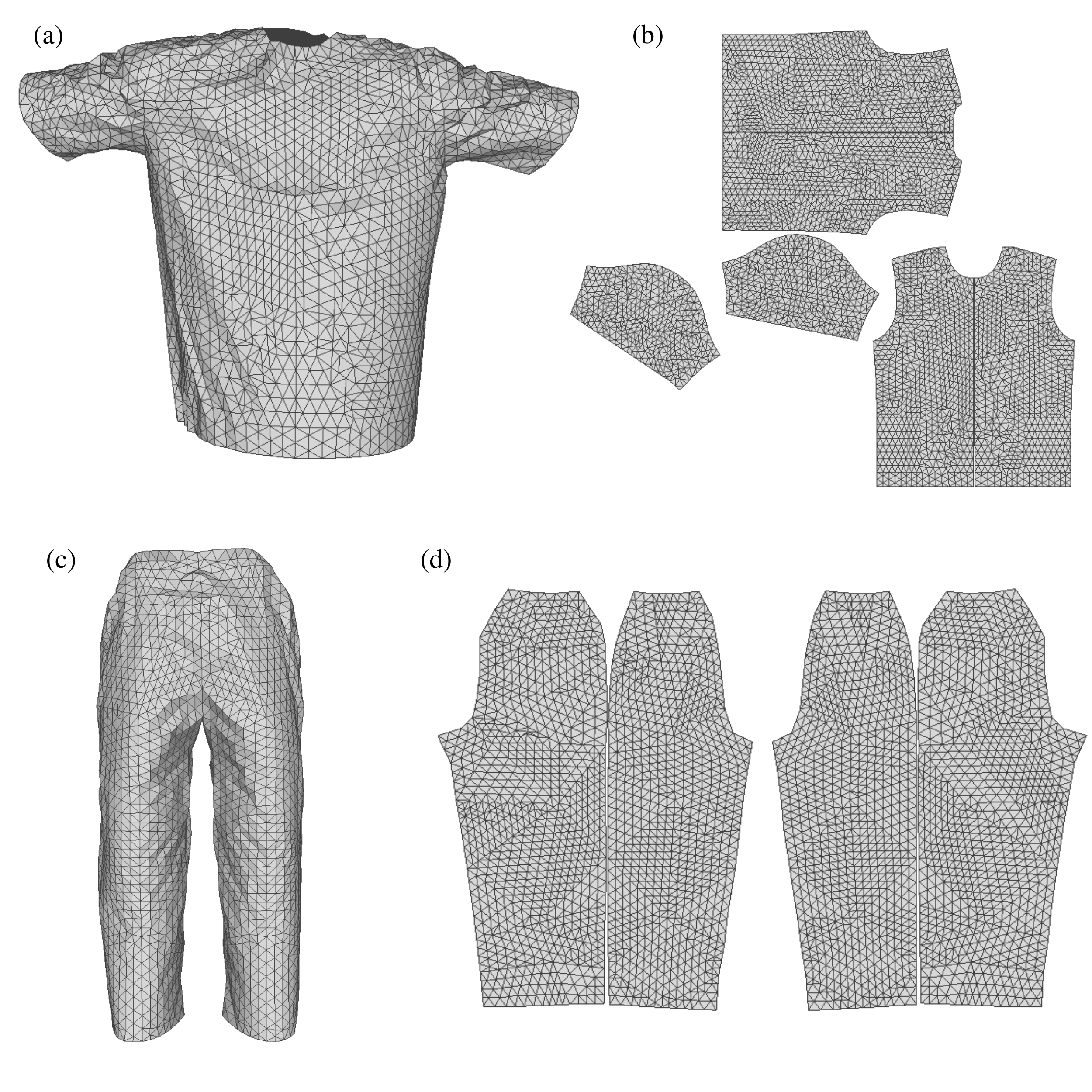}
   \caption{Our approach uses garment templates for modeling and texturing. (a) The template mesh for T-shirt, whose texture coordinates match the vertex coordinates of the (b) reference mesh. (c) Template mesh for pants, and the corresponding (d) reference mesh.}
\label{fig_garmenttemplate}
\end{center}
\end{figure}

We use 3D garment models from Berkeley Garment Libraries \cite{BerkeleyGarmentLib} as templates. For each garment type, a coarse base mesh and a finer isotropically refined mesh are provided by the library. We use the refined mesh in world-space configuration as our base model. In addition, the texture coordinates of the refined mesh store the material coordinates that refer to a planar reference mesh. We use this 2D reference mesh for texture extraction. Currently, our system supports two garment types: T-shirt and pants as shown in Fig. \ref{fig_garmenttemplate}.

\subsubsection{3D Model Deformation}
To create 3D garment models that conform to the sizing information from the input images, we apply Free-Form Deformation (FFD) \cite{Sederberg:1986:FDS} to deform a garment template. We chose FFD because it can be applied to 3D models locally while maintaining derivative continuity with adjacent regions of the model. For two view data (front and back), FFD is a plausible solution. When there are multi-view images, videos, or 4D scans of garments, other mesh fitting techniques can be used to generate more accurate results.

\begin{table}[b!]
  \caption{Control Points Distances from Landmarks for T-shirt}
  \label{tab_tshirt}
  \scriptsize%
	\centering%
  \begin{tabu}{p{0.3\linewidth} p{0.6\linewidth}}
  \toprule
   \textbf{Distance} & \textbf{How to calculate}  \\
  \midrule
	$D(P\textsubscript{0jk},\ P\textsubscript{1jk})$ & left sleeve length * $cos(\alpha)$   \\
	$D(P\textsubscript{1jk},\ P\textsubscript{2jk})$ & chest width (armpit\_left to armpit\_right)\\
	$D(P\textsubscript{2jk},\ P\textsubscript{3jk})$ & 
	right sleeve length * $cos(\beta)$  \\
	$D(P\textsubscript{ij0},\ P\textsubscript{ij1})$ &
    distance from armpit to hemline   \\
    $D(P\textsubscript{ij1},\ P\textsubscript{ij2})$ & 
    distance from armpit to shoulder \\
    $D(P\textsubscript{ij0},\ P\textsubscript{ij3})$ &     
	distance from neck to hemline\\
	$D(P\textsubscript{i0k},\ P\textsubscript{i1k})$ &     
	$D(P\textsubscript{ij1},\ P\textsubscript{ij2})$ * $S$\\
	$S$ & $D(P\textsubscript{i0k},\ P\textsubscript{i1k}) / D(P\textsubscript{ij1},\ P\textsubscript{ij2})$, un-displaced. \\
  \bottomrule
  
  \end{tabu}%
\end{table}

\begin{figure}[t!]
\begin{center}
   \includegraphics[width=0.98\linewidth]{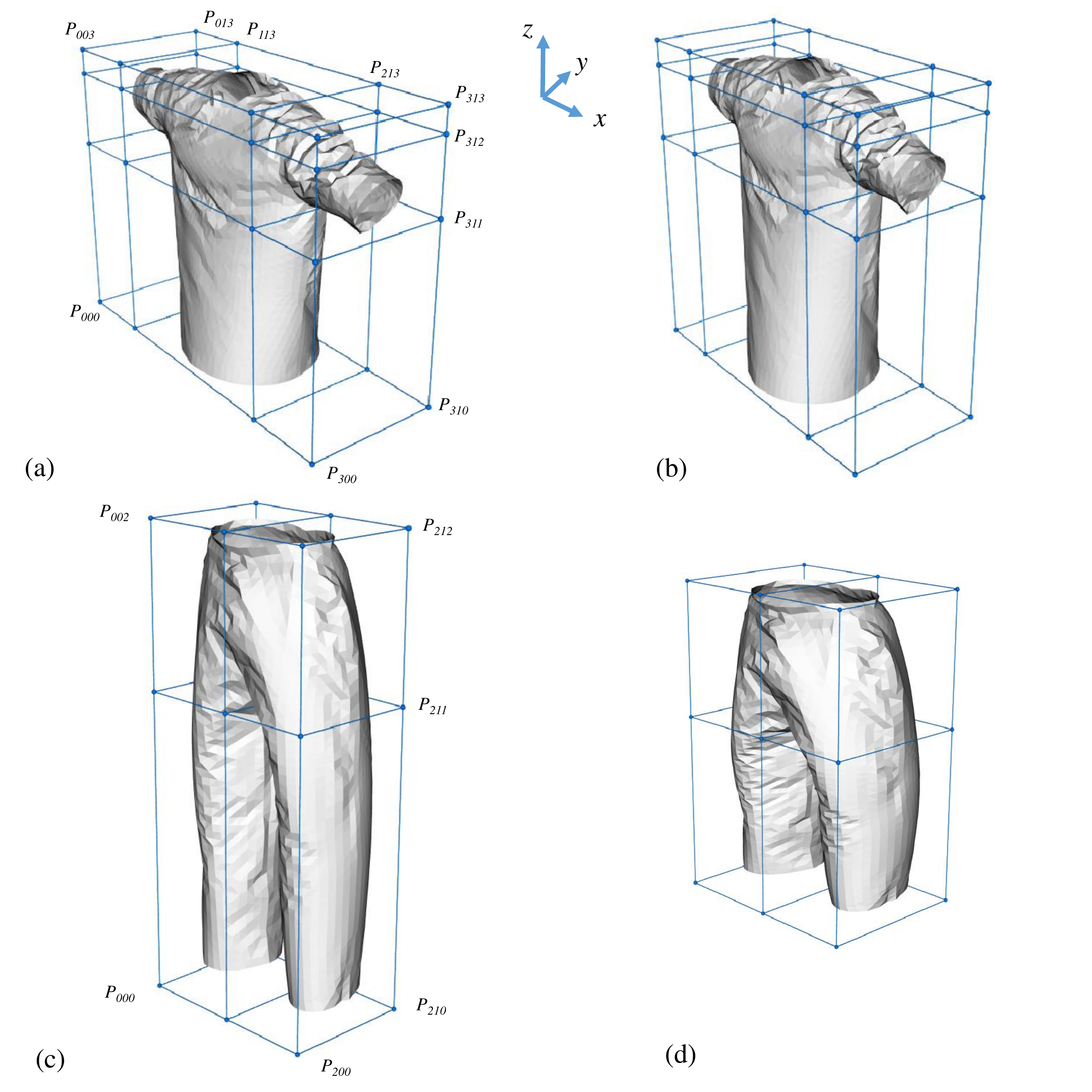}
   \caption{ Template Deformation. (a) The original template for T-shirt with control grid. (b) Deformed template that captures a different shape. (c) The original template for pants. (b) Deformed template.}
\label{fig_ffd}
\end{center}
\end{figure}

For each garment template, we impose a grid of control points ${P\textsubscript{ijk}}\ (0\leq{i}<l,\ 0\leq{j}<m,\ 0\leq{k}<n)$ on a lattice. The deformation of the template is achieved by moving each control point ${P\textsubscript{ijk}}$ from its original position. Control points are carefully chosen to facilitate deformation of individual parts so that a variety of garment shapes can be modeled. For T-shirt, as shown in Fig. \ref{fig_ffd} (a, b), we use $l=4, \ m=2, \ n=4$. For pants, as shown in Fig. \ref{fig_ffd} (c, d), we use control points with $l=3, \ m=2, \ n=3$.
 
If metric scale of the resulting 3D model is desired, we ask the user to specify a measurement \textit{l} in world space (e.g., sleeve length). Otherwise, a default value is assigned to \textit{l}. Based on the ratio between image space sleeve length to \textit{l}, we can convert any image space distance to world space distance.

\begin{table}[b!]
  \caption{Computing Control Points Distances for Pants}
  \label{tab_pants}
  \scriptsize%
	\centering%
  \begin{tabu}{p{0.3\linewidth} p{0.6\linewidth}}
  \toprule
   \textbf{Control Points} & \textbf{How to calculate}  \\
  \midrule
	$D(P\textsubscript{0jk},\ P\textsubscript{1jk})$ & un-displaced distance * $S^*$\\
	$D(P\textsubscript{1jk},\ P\textsubscript{2jk})$ &  un-displaced distance * $S^*$\\

	$D(P\textsubscript{ij0},\ P\textsubscript{ij1})$ &
    distance from crotch to bottom   \\
    $D(P\textsubscript{ij1},\ P\textsubscript{ij2})$ & 
    distance from crotch to waist line  \\
	$D(P\textsubscript{i0k},\ P\textsubscript{i1k})$ &  un-displaced distance * $S^*$\\

  \bottomrule
  \multicolumn{2}{l}{$^* S$ is ratio between new waist girth to template waist girth.} \\

  \end{tabu}%
\end{table}

FFD control points do not directly corresponded to image landmarks. Instead, we compute 2D distances between garment landmarks and use them to compute 3D distances between control points.  
Tab. \ref{tab_tshirt} shows how to calculate control point distances for the T-shirt type.  Constants $alpha$ and $beta$ are the angle between horizontal direction and left sleeve and the angle between horizontal direction and right sleeve respectively. They are measured from the template T-shirt mesh. The distances are then used to compute new locations of control points for template mesh deformation. 

Since the T-shirt template resembles the shape of a T-shirt on a mannequin, using photos of T-shirts on mannequins achieves most accurate results. On such images, the distance between two armpits corresponds to the chest width of the mannequin. When a T-shirt lays on a flat surface, the  distance between two armpits corresponds to half perimeter of the chest.  In this case, we fit an ellipse to the horizontal section of the chest. We then compute the width of the horizontal section as the major axis of the ellipse using the perimeter measurement. Images of fashion models are not suitable for garment size
estimation due to self-occlusion, wrinkles, etc. Tab. \ref{tab_pants} shows the calculation of control points for the pants.

\subsection{Texture Extraction}
The texture coordinates in the 3D mesh refer to the vertices in the planar 2D reference mesh. This allows us to perform 3D texture mapping by mapping input images onto the 2D reference mesh as a surrogate. The different pieces in the reference mesh correspond to different garment segmentation parts. This is the reason semantic segmentation is performed during garment image analysis. Texture mapping becomes an image deformation problem where the source is a garment part (e.g., left sleeve) and the target is its corresponding piece on the reference mesh. 

\begin{figure}[b!]
\begin{center}
   \includegraphics[width=0.99\linewidth]{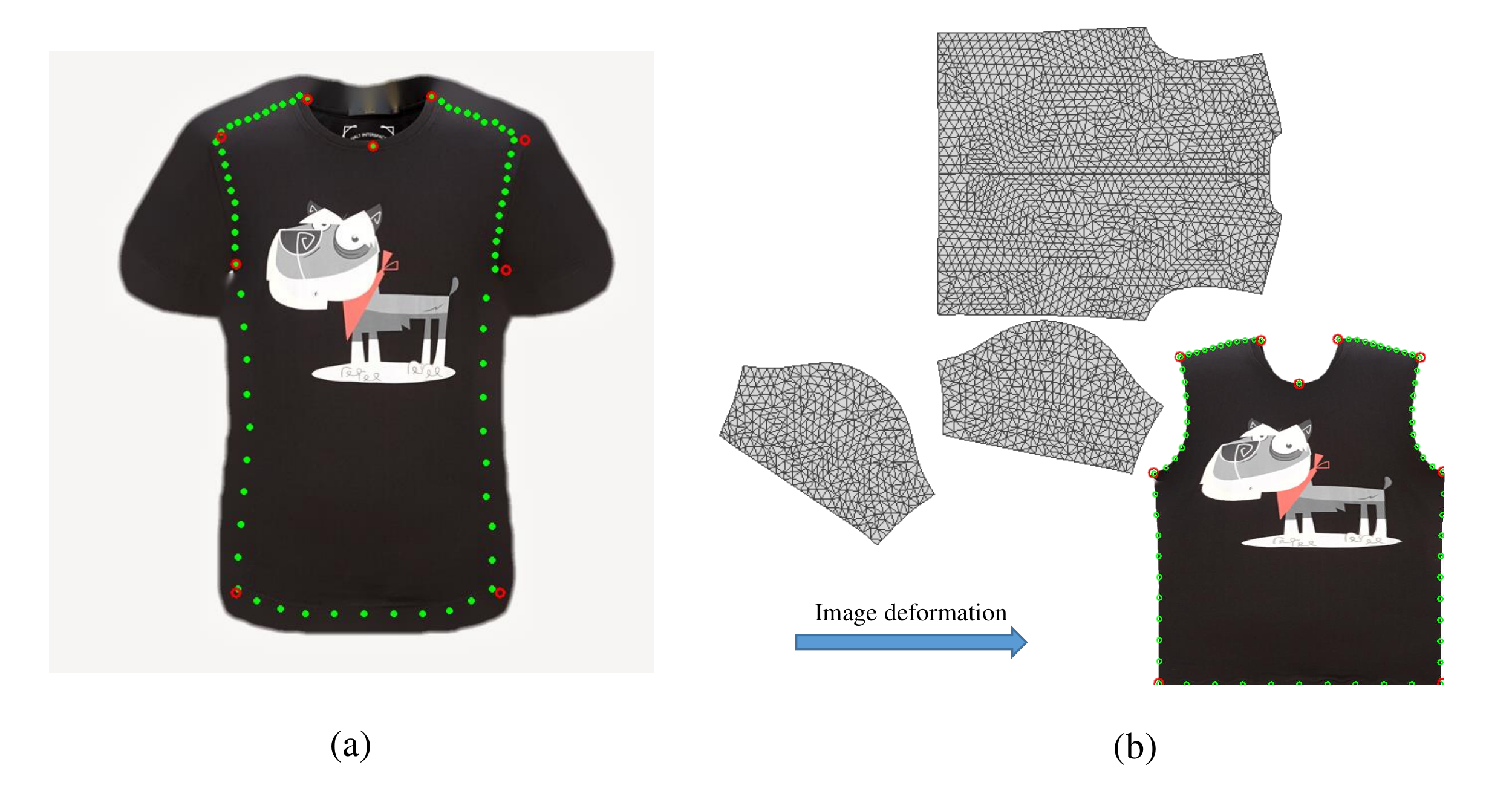}
   \caption{ Texture Extraction for T-Shirt. (a) The extrapolated T-shirt image with control points computed along the contour of the front segment. (b) The front segment is deformed to the desired location on the 2D reference mesh.}
\label{fig_texture}
\end{center}
\end{figure}

On the reference mesh, we manually label the landmarks (Fig. \ref{fig_texture} (b) red circles). This only needs to be done once for each garment type. In this way, we establish feature correspondence between predicted landmarks on the source image and manually-labeled landmarks on the target image. However, using a sparse set of control points leads to large local deformation, especially around contours. To mitigate this, we map each landmark point onto the contour of the part by finding the closest point on the part contour. Then between each pair of adjacent landmarks, we sample \textit{N} additional points uniformly along the contour. We do this for both input garment image and reference mesh (green circles in Fig. \ref{fig_texture}).The corresponding points are then used by Moving Least Squares (MLS) method with similarity deformation  \cite{Schaefer:2006:IDU:1141911.1141920} to deform textures from the input image to the reference mesh. Alternatively, a Thin Plate Spline (TPS) based approach similar to that used in VITON \cite{VITON_2018} can also be used for image warping.

\begin{figure}[h!]
\begin{center}
   \includegraphics[width=0.99\linewidth]{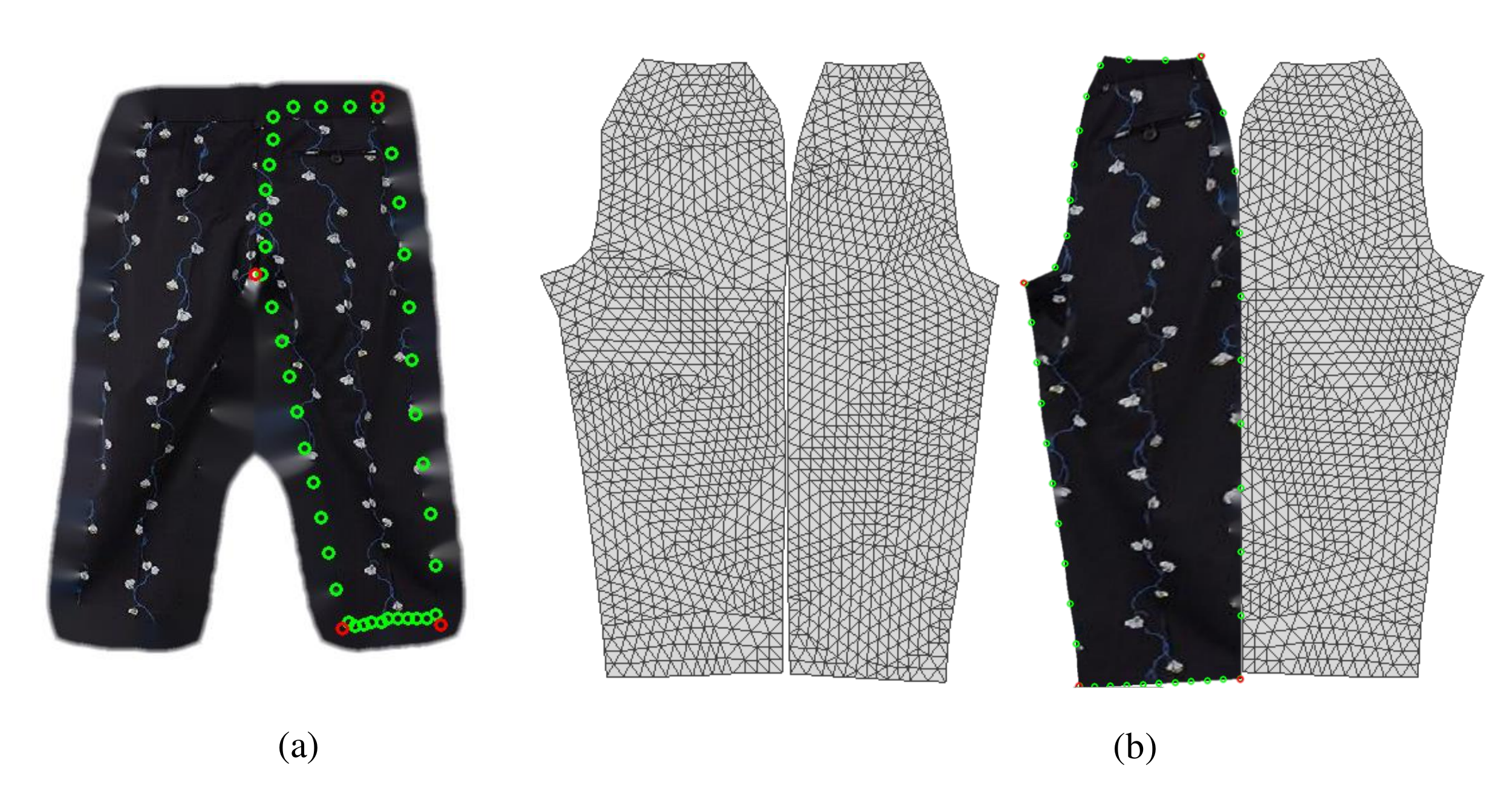}
   \caption{ Texture Extraction for Pants. (a) The extrapolated pants image with control points. (b) The image segment is deformed to the desired location on the 2D reference mesh.}
\label{fig_texture_pants}
\end{center}
\end{figure}

Before image deformation, each garment segment is eroded slightly to accommodate for segmentation artifacts. Then, color texture is extrapolated from the garment to surrounding area to remove background color after deformation. Fig. \ref{fig_texture} shows the process of deforming the front segment of a T-shirt to the desired location on its 2D reference mesh. Fig. \ref{fig_texture_pants} shows that for the right leg of pants. Note that to better illustrate the idea, we use a small value of $N=10$ in Fig.\ref{fig_texture} and \ref{fig_texture_pants}. In our experiments, we found that denser control point set (e.g. $N=50$) works better.

In our current implementation, the back piece around the neck/collar is often included in the front piece segmentation result. To handle this, we cut out the back piece automatically. JFNet predicts the front middle point of the neck as a landmark. We then correct the front piece segmentation by tracing the edge from two shoulder points to the middle neck point.


\section{Experiments}
In this section, we show quantitative experimental results for JFNet. We also show results on 3D modeling.

\subsection{Evaluation of JFNet}

Our model requires both landmark and segmentation annotations, thus we cannot compare our results directly with other SOTAs by training our model on public dataset. Nevertheless, we have trained CPM and DeepLabV3+ on our dataset and compare them with JFNet.

We trained JFNet for tops and pants separately. For each model, 2,000 images are used for training and 500 images for validation. Evaluation is performed on the remaining 500 images. We used the standard intersection over union ($IoU$) criterion and mean $IoU$ ($mIOU$) accuracy for segmentation evaluation and normalized error $(NE)$
metric\cite{Liu:2016:fashionlandmark} for landmark prediction evaluation. $NE$ refers to the distance between predicted landmarks and ground truth locations in the normalized coordinate space ($i.e.$, normalized with respect to the width of the image), and it is a commonly used evaluation metric.

Tab. \ref{2dquant_res} shows performances of different methods. For both tops and pants, JFNet achieves better performance on \textit{both} landmark prediction and garment part segmentation. Our landmark prediction on tops greatly
outperforms CPM (0.031 vs. 0.075). This shows that constraints and guidance from segmentation task have helped landmark prediction. Landmark prediction performance on pants also improves, but not as much because landmarks of pants are less complex than those of tops. Part segmentation is a more complex task. Thus, it is reasonable that our model does not boost the segmentation task as much. Nevertheless, JFNet still improves upon DeepLabV3+. 

It is worth noting that the purpose of the proposed model is to handle multiple tasks simultaneously with performance improvement compared to individual tasks. Thus, our method focuses on information sharing and multi-task training while other SOTAs focus on network structure and training for each individual task. In the future, we can also incorporate other SOTA networks into our joint learning model.

\begin{table}[h]
\begin{center}
\caption{Landmark Prediction and Garment Segmentation Performance Comparison}
\label{2dquant_res} 
\begin{tabular}{ccccccccc}
\hline  \hline
&\multicolumn{2}{c}{} & \multicolumn{2}{c}{Tops} & \multicolumn{2}{c}{} &\multicolumn{2}{c}{Pants}\\ [0.5ex]
  \cline{4-5}  \cline{8-9}
Methods & \multicolumn{2}{c}{} &  NE & mIOU & \multicolumn{2}{c}{} & NE & mIOU\\ [0.5ex]
  \hline
$CPM$ \cite{Wei:2016:CPM} & \multicolumn{2}{c}{} & $0.075$ & $-$ & \multicolumn{2}{c}{} & $0.034$ & $-$\\ 
$Deeplabv3+$ \cite{Chen:2018:DEEPLABV3+} & \multicolumn{2}{c}{} & $-$ & $0.721$ & \multicolumn{2}{c}{} & $-$ & $0.964$\\ 

  \hline
$JFNet$ & \multicolumn{2}{c}{} & $\textbf{0.031}$  & $\textbf{0.725}$ & \multicolumn{2}{c}{} & $\textbf{0.022}$ & $\textbf{0.968}$\\ 

  \hline
 \end{tabular}
\end{center}

\end{table}

\subsection{3D Modeling Results}

\begin{figure*}[h!]
\begin{center}
   \includegraphics[width=0.95\linewidth]{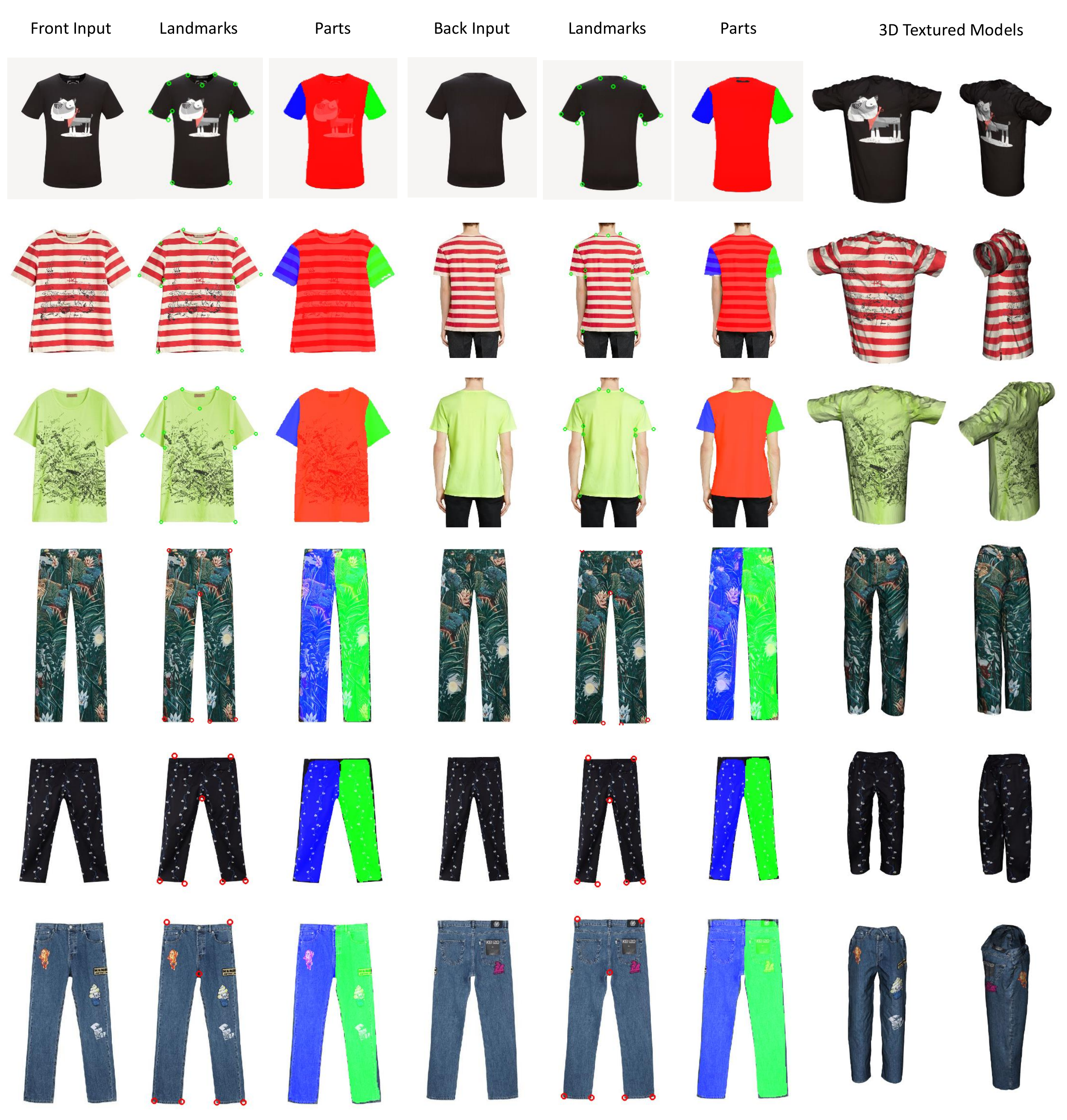}

   \caption{3D Modeling Results. On each row we show front image and its landmark prediction and part segmentation, followed by back image and its landmark and part segmentation results. The final two columns show 3D textured models for two view points.}
\label{fig_result}
\end{center}
\end{figure*}

We applied our 3D garment modeling algorithm on various input images and the results are in Fig. \ref{fig_result}. Our approach utilizes the sizing information estimated from fashion landmarks to model different styles of garments (e.g., different length of legs or different fits of T-shirt).  For example, the 3rd T-shirt is slightly longer, the 2nd T-shirt is slight wider, and the 1st T-shirt has narrower sleeves. These correspond to the characteristics of the input garment images. Our approach can also extract textures from garment images and map them on to different parts of the constructed 3D model. 

To quantitatively evaluate our 3D modeling is expensive. This involves capturing 2D images of various garments
and scanning them into 3D models. An alternative is to use synthetic data with ground truth to evaluate accuracy of size estimation and 3D reconstruction. We leave these for future work. Nevertheless, 3D modeling results of our approach are visually plausible for applications where accuracy requirement is not strict.

\section{Conclusion}
We present a complete system that takes photos of a garment as input and creates a 3D textured virtual model. We propose a multi-task network called JFNet to predict fashion landmarks and segment the garment into parts. The landmark prediction results are used to guide template-based deformation. The semantic part segmentation results are used for texture extraction. We show that our system can create 3D virtual models for T-shirt and pants effectively.

\section{Limitation}
One limitation is due to the representation power of the templates. Because our model is deformed from a template, the shape of the template limits the range of garments we can model. For example, our pants template is a regular fit. Modeling slim or skinny pants will be impractical. Our approach recovers shape, but not the pose of the garment. To learn the 3D pose of garments, more data and annotations are required.

Another limitation is that we only use two photos (front and back view) for texture extraction. This leads to excessive local deformation when source and target contours are very different (see stickers on the jeans in Fig. \ref{fig_result} last row).

The photo sets for testing our 3D modeling approach are from online shopping sites. Two occlusion-free images can always be selected from each set. In general, occlusion can pose a problem for texture extraction. However, missing textures can be mitigated using image in-painting. Missing landmarks can be mitigated using symmetry-based landmark completion.

Finally, our system only supports T-shirt and pants now and we only address a simplified version of the garment modeling problem, which usually involves wrinkles, folds and pleats.

\section{Future Work}
Currently, 2D proportions from the photos are transferred to the 3D model. In the future, We want to use a garment modeling approach that uses sewing patterns \cite{Jeong:2015:doi:10.1002/cav.1653}. We can fit the shape of each individual 2D sewing pattern using image part segmentation. Then, these 2D patterns can be assembled in 3D space as in commercial garment design process. In this way, we can better transfer the shapes from 2D images to 3D models.

We also want to investigate if more than two images can be used together to texture a 3D model \cite{Detail_human_avartar_2018}. The distorted textures along the silhouettes of front and back view can be filled in by a side view photo.

For applications that require accurate 3D information, we would like to perform quantitatively evaluation of our 3D modeling algorithm. 

Finally, by incorporating more garment templates, more garment types can be supported. Since we only need to create a template once for each type/fit, the overhead is small if used in large scales. There are certain garments that are not suitable for our approach (e.g., fancy dresses with customized design). A possible approach is to use a hybrid system where template-based deformation generates a base model and 3D details can be added via other methods. Part segmentation in its current state is not suitable for open jackets. It would be interesting to see if semantic segmentation model with more data and annotation can distinguish between back side and front side.

\acknowledgments{The authors wish to thank the reviewers for their insightful comments and suggestions.}

\bibliographystyle{abbrv-doi}

\bibliography{reference}
\end{document}